# Evaluation of GPT-4o and GPT-4o-mini's Vision Capabilities for Compositional Analysis from Dried Solution Drops

Deven B. Dangi[1], Beni B. Dangi[2,*], and Oliver Steinbock[1,*]

1) Florida State University, Department of Chemistry and Biochemistry, Tallahassee, FL 32306-4390, USA

2) Florida Agricultural and Mechanical University, Department of Chemistry, Tallahassee, FL 32307, USA

*) Email: beni.dangi@famu.edu, osteinbock.fsu.edu

## Abstract

When microliter drops of salt solutions dry on non-porous surfaces, they form erratic yet characteristic deposit patterns influenced by complex crystallization dynamics and fluid motion. Using OpenAI's image-enabled language models, we analyzed deposits from 12 salts with 200 images per salt and per model. GPT-4o classified 57% of the salts accurately, significantly outperforming random chance and GPT-4o mini. This study underscores the promise of general-use AI tools for reliably identifying salts from their drying patterns.

## ToC Graphics

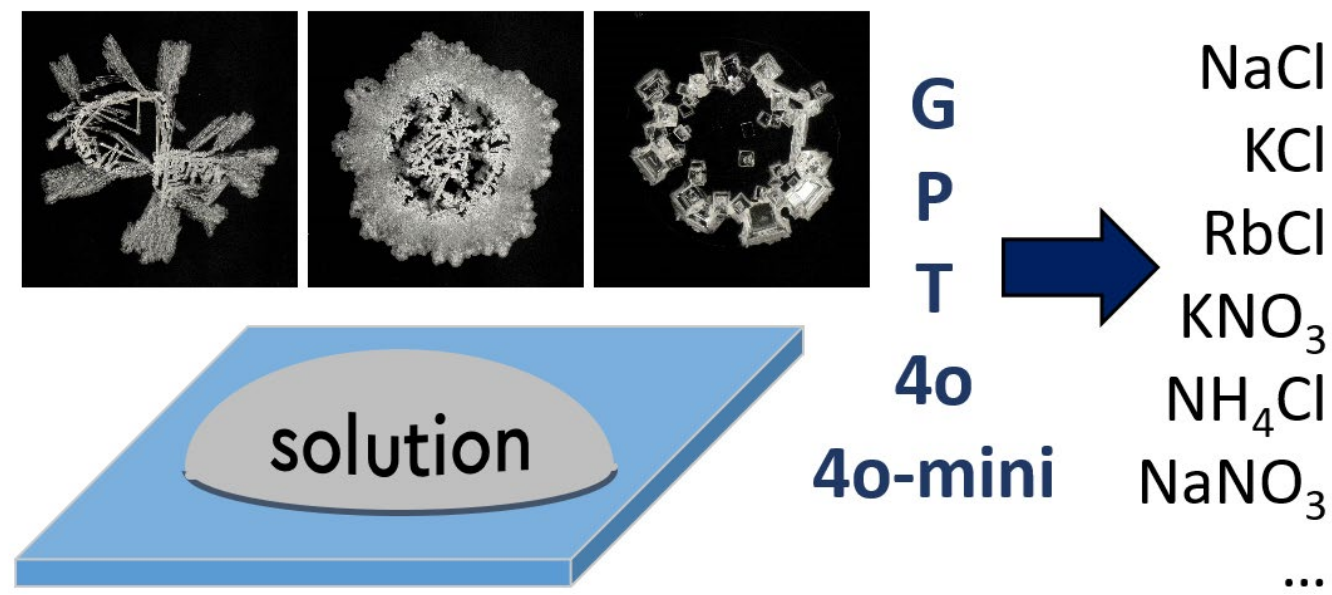

## 1. Introduction

The macroscopic patterns and compositional variations formed by drying solutions have long fascinated scientists. For instance, chemistry's first Nobel laureate, van 't Hoff, aimed to predict the mineral sequence of evaporating seawater, partly motivated by economically important potash deposits.[1] Another example is the creeping salt phenomenon, where dissolved salts migrate along surfaces during evaporation, leading to distinctive deposition patterns influenced by capillary forces and crystallization dynamics.[2]

A specific variant of this type of pattern formation are the deposit structures formed by sessile evaporating drops of solutions and dispersions. An iconic case is the coffee-ring effect, where particles suspended in an evaporating droplet migrate to the periphery, leaving a characteristic ring-like deposit.[3] Lesser-known examples include self-lifting macroscopic NaCl crystals, where smaller crystals grow into "legs" that push larger crystals upward from a hydrophobic surface during the final stages of evaporation.[4]

The diverse dynamics of deposit formation have inspired studies on using drying patterns to reveal liquid compositions. For example, deposits from tap water and alcoholic beverages have been analyzed for compositional features. In diagnostics, fern-like patterns in dried tears have been linked to dry-eye disease[5], while dried blood patterns show potential for detecting leukemia and anemia.[6] Similarly, deposits of KCl or KCl/$MgCl_2$ mixed with urine have been analyzed with deep neural networks to assist in diagnosing bladder cancer.[7]

Our groups have studied the deposit patterns formed by 10 μL salt solutions on glass. The first study, analyzed twelve inorganic salts based on a total of 6,000 images using a reduction of each photo to 16 image metrics.[8] Simple optimization of Euclidian distances in the resulting 16-dimensional morphospace yielded prediction accuracies of 90% from single test images. Machine learning analyses of 42 different salts and only 14 training images, still yielded correct salt identification in 75% of all cases.

Very recently, we developed an automated experimental method for the rapid collection of large image libraries of deposit patterns.[9] This robotic drop imager (RODI) allowed the analysis of seven different salts at five different initial concentrations. Using a deep learning method, higher-dimensional metrics vectors, and over 23,000 images, we achieved prediction accuracies of 98.6% for the salt type and 92.5% for the combined salt type and initial concentration.

Here, we investigate a subset of the image data in ref. 8 using a very different approach: multimodal large language models (LLMs). Unlike traditional image analysis techniques, these models combine advanced natural language processing capabilities with image recognition by integrating visual inputs into their neural network architecture. Images are processed as input, allowing the models to extract and interpret features, which are then correlated with textual or structured data through a shared embedding space. [10] Recent work on GPT-4's vision capabilities has demonstrated success in tasks such as interpreting complex visual scenes and analyzing graphs, suggesting its potential for non-standard applications like salt deposit analysis.[11,12]

## 2. Methods

The GPT-4o and GPT-4o-mini models exhibit broad general knowledge but lack the specific training to recognize macroscopic images of salt deposits formed by evaporating droplets.[13] To address this, a curated set of training images was included to familiarize the models with the characteristic patterns of each salt type. These images, generated in a laboratory setting, have been described previously.[8] All images and the used Python script are available for download at refs. 14 and 15, respectively.

OpenAI's Application Programming Interfaces (APIs) were employed to facilitate batch processing, enabling multiple simultaneous requests with higher token limits and reduced costs compared to sequential approaches.[16,17] Each request was stateless—no memory was retained between interactions[18]—therefore, the training images required for salt differentiation had to be included in every individual request. While this ensured consistent processing, it also substantially increased the token count, contributing to the overall computational cost.

For batch processing, a Javascript Object Notation Line (JSONL) file was prepared, with each line containing a valid JSON object representing one API request. Because of file size and token constraints, each trial required multiple JSONL files. Specifically, 200 images from each salt were randomly selected to capture morphological diversity; these images were then added to a JSONL file as individual requests. Consequently, each trial comprised 2,400 requests (200 images per salt for 12 salts), distributed across as many JSONL files as necessary. To reduce image repetition, we split the 200 images per salt into two sets of 100 images that contained no duplicates.

Each request contained several key parameters: (1) an ID combining the image's filename (a randomly generated unique number) with its parent folder name, allowing later confirmation of the true salt identity; (2) the specified model name (either gpt-4o-mini-2024-07-18 or gpt-4o-2024-08-06); and (3) a temperature set to 0 and a seed set to 17 (arbitrarily chosen) to bring results closer to being deterministic, despite the inherently probabilistic nature of these models.[19] Additionally, each request included the following system prompt:

***"You are a helpful assistant who is knowledgeable about different types of salt crystals and can identify them from images. You can identify these 12 different salts: NaCl, KCl, NH4Cl, Na2SO4, K2SO4, NH4NO3, NaH2PO4, NaNO3, Na3PO4, KBr, KNO3, and RbCl."***

For training, each request was accompanied by 12 user messages passed as context, each containing five images and a line of text informing the model of the training images' identities. These images were compressed and contained in Base64 format in every request (see Python script). A 13th user message stating,

***"Identify this salt with just the name."***,

was then included alongside the test image, prompting the model to provide only a single-salt name as its output.

Once all JSONL files were programmatically created, they were submitted to the designated model. For convenience, the OpenAI web interface[20] was utilized, offering the same functionality as the API. Initially, usage tier restrictions limited file size and permitted only one smaller file to be processed at a time; however, elevated usage tiers allowed multiple large files to be processed concurrently, circumventing rate limits. Each fully processed file produced an output JSONL file containing the response for every request in the input file. To analyze these results, the custom ID was used to match each image back to its correct salt type, and the first mention (typically the only one) of any of the 12 salts in the model's output was recorded as the prediction.

## 3. Results and Discussion

Examples for the deposit patterns formed during the evaporation of the twelve saturated salt solutions are shown in Figure 1. Some of the salts form rather characteristic patterns, such as the $NH_4NO_3$ which is a creeping salt or NaCl and KCl which dry into small crystals. However, the variations of the deposit patterns formed by any given salt can be substantial and as discussed in ref. 8 some compounds even form two distinct pattern types, a phenomenon that was coined "bifurcated salts".

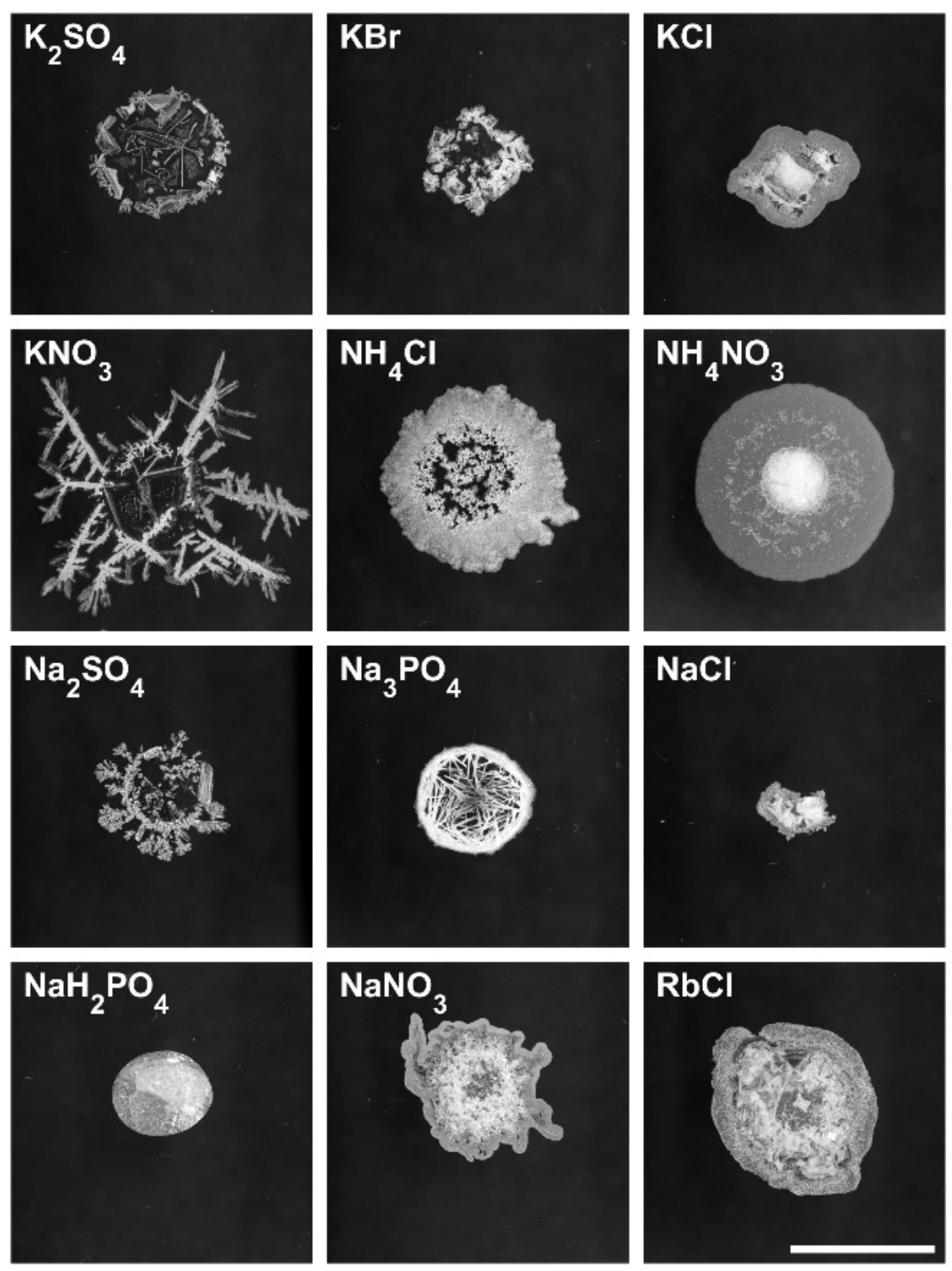

**Figure 1.** Evaporating solution drops create deposit patterns that depend on the dissolved salt. The figure shows examples for the 12 analyzed salts taken from the training sets with enhanced contrast. Scale bar: 1 cm (applies to all panels).

Our attempts to identify the salt type using GPT 4o-mini and 4o used photos similar to those in Fig. 1. As detailed in the Methods section, each test was performed by providing five training images for each of the 12 salts and then testing the response with 200 random images for each salt and model. We reemphasize that the key steps of these identification attempts were plain English prompts.

Figure 2 shows the main results of our study as confusion matrices. The ordinate denotes the true salt name, while the abscissa shows the predicted chemical formulas. Accordingly, high numbers along the top-left to bottom-right diagonal indicate successful predictions. Notice that the numbers in each row add up to 200, i.e. the total number of images presented for identification.

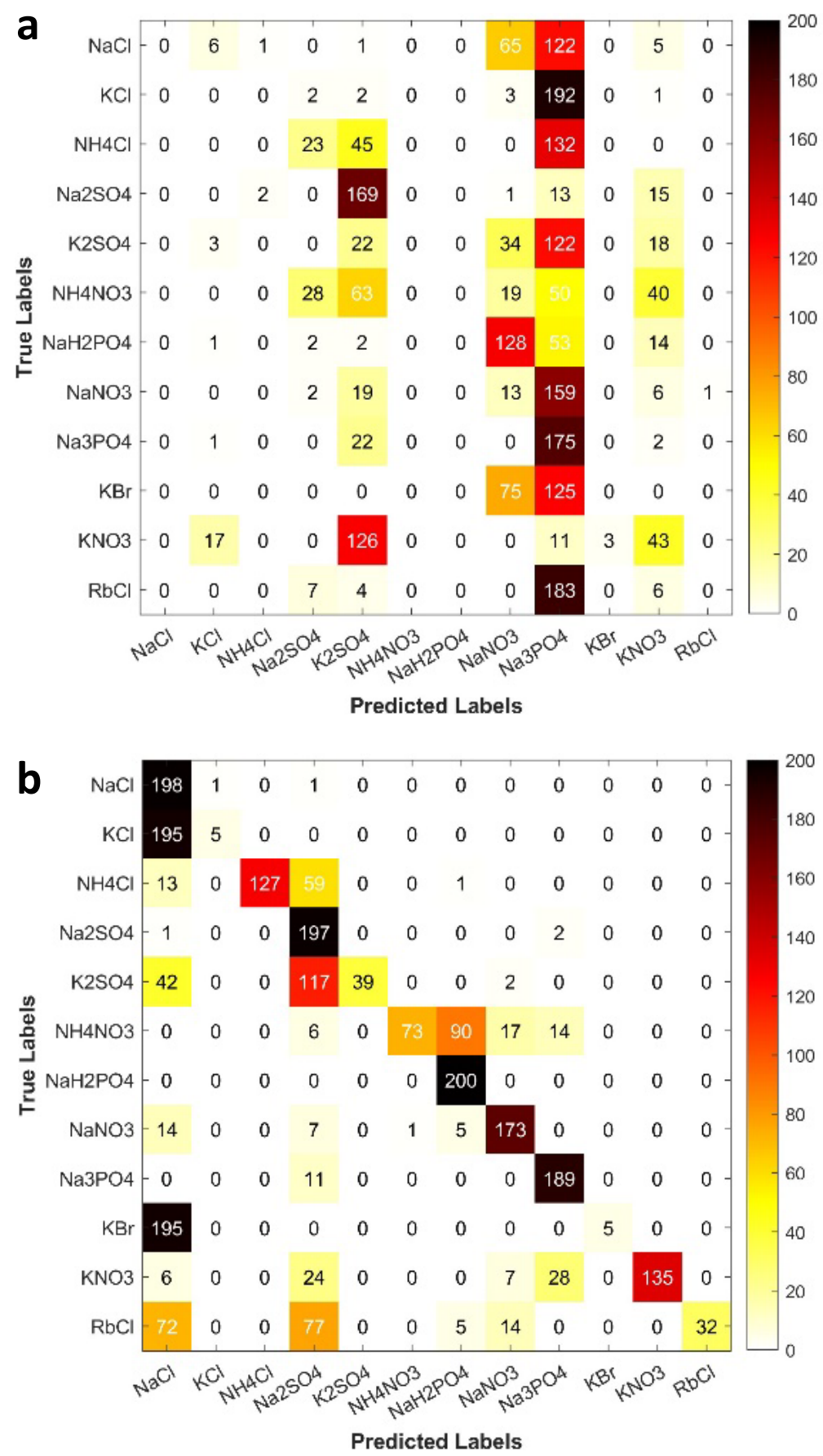

**Figure 2.** Confusion matrices summarizing the prediction outcomes of a) GPT 4o-mini and b) GPT 4o. The corresponding total prediction accuracies are a) 10.5 % and b) 57.2 %.

The performance of the GPT 4o-mini model is summarized in Fig. 2a. The results are overall disappointing with a total prediction accuracy of 10.5 % which is only marginally above the expectation of 8.3 % for random guesses. Furthermore, we find that the model shows a peculiar preference for identifying patterns as $Na_3PO_4$ and to a lesser extent as $NaNO_3$. Indeed, 55.7% of all queries are identified as $Na_3PO_4$, which in the first prompt's list of salt occupies the unremarkable tenth position. The origin of this preference is unknown.

Despite the poor performance of the mini model, GPT 4o demonstrates a clear ability to carry out the image-based identification task. The confusion matrix for GPT 4o in Fig. 2b shows large counts along the diagonal and corresponds to an overall prediction accuracy of 57.2 %. Notice that the score for $NaH_2PO_4$ is a perfect 200 out of 200. The weakest result is found for KBr and KCl with correct scores of 5 out of 200 and typical misidentifications as NaCl. We note that visual inspection of the latter three salt patterns indicates close similarities (see Fig. 1).

We also investigated whether the repeated runs for each model suggest statistically significant differences. For this analysis, we determined Cohen's kappa coefficients (Tab. S1), a measure of inter-rater agreement that corrects for chance, thus allowing us to assess how consistently each model performed across multiple trials. For the two GPT 4o sets, we found $\kappa = 0.962$, indicating near-perfect agreement between the repeated runs and underscoring the model's robust reproducibility. Similarly, the GPT 4o-mini sets yielded $\kappa = 0.907$.

Lastly, we computed F1 scores (Tab. S2), a single metric that balances precision and recall, offering a deeper understanding of how the models handle each salt classification under imbalanced conditions. The macro F1 scores for each model differed slightly from their overall accuracies, indicating that even with equal representation of all salts in our training set, certain salts pose more challenges. Notably, both GPT 4o and GPT 4o-mini had lower F1 scores for visually similar salts such as KBr, KCl, and NaCl. Nevertheless, GPT 4o consistently outperformed GPT 4o-mini, highlighting the enhanced capability of the larger model in this classification task.

**4. Conclusions**

Artificial intelligence tools offer unprecedented opportunities for novel chemical and biological analyses from image data.[21] In this context, our team recently showed that specialized approaches such as multi-layer perceptrons and other deep learning models can achieve accuracies of > 90% for the identification of inorganic salts from macroscopic deposit patterns formed during the evaporation of solution drops.[8,9] Here we studied how well multi-modal LLMs like GPT 4o can perform similar tasks. Our results indicate clear promise in such a non-tailored approach. We also showed that performance is strongly model dependent with GPT 4o-mini barely outperforming random guesses. Considering the rapid rate of improvement of multi-modal LLMs, it seems likely that at least some image analysis tasks in chemistry will soon be solvable with broadly accessible tools. As these general-use models continue to mature, they may no longer require specialized architectures or extensive domain-specific datasets to achieve high accuracy, ultimately expanding the reach and impact of AI-driven chemical analyses.

## ASSOCIATED CONTENT

### Data Availability Statement

The Python script used for generating JSONL files for submission into OpenAI's API, training images, csv files of the data in Fig. 2, and figure-generating MATLAB scripts are freely available at GitHub: https://github.com/osteinbock/GPT4oBatch

All image data is freely available at https://www.chem.fsu.edu/~steinbock/saltscapes.php

### Supporting Information

The Supporting Information is available free of charge at https://pubs.acs.org/doi/...

Cohen's kappa values between pairs of analysis sets; F1 scores for every salt across all trials.

## AUTHOR INFORMATION

### Author contributions

D.B.D. and O.S. designed research; D.B.D. performed research and analyzed data; B.B.D. and O.S. funding acquisition; and D.B.D., B.B.D., and O.S. wrote the paper.

### Funding

This material is based on work supported by NASA under grant no. 80NSSC23M0050.

### Notes

The authors declare no competing financial interest.

## ACKNOWLEDGMENTS

We thank Dr. Bruno C. Batista, Dr. Amrutha S V, Dr. Jéssica A. Nogueira, Prof. Jie Yan, Dr. Suman Sinha Ray, Dr. Ruth Agada, and Srinivasakranthikiran Kolachina for discussions.

**SUPPORTING INFORMATION**

**Evaluation of GPT-4o and GPT-4o-mini's Vision Capabilities for Compositional Analysis from Dried Solution Drops**

Deven B. Dangi[1], Beni B. Dangi[2,*], and Oliver Steinbock[1,*]

1) Florida State University, Department of Chemistry and Biochemistry, Tallahassee, FL 32306-4390, USA

2) Florida Agricultural and Mechanical University, Department of Chemistry, Tallahassee, FL 32307, USA

*) Email: beni.dangi@famu.edu, osteinbock.fsu.edu

**Table S1.** Cohen's Kappa Values. These statistical measures assess the degree of agreement between two methods, beyond what would be expected by chance. Values close to 1.0 signify agreement, whereas values close to 0.0 indicate random chance.

| | Set 1 (4o-mini) | Set 2 (4o-mini) | Set 1 (4o) | Set 2 (4o) |
|---|---|---|---|---|
| Set 1 (4o-mini) | 1 | 0.9073 | 0.0609 | 0.0645 |
| Set 2 (4o-mini) | 0.9073 | 1 | 0.06 | 0.0647 |
| Set 1 (4o) | 0609 | 0.06 | 1 | 0.9618 |
| Set 2 (4o) | 0.0645 | 0.0647 | 0.9618 | 1 |

**Table S2.** F1 Scores by Salt Type and Model.

| | Set 1 (4o-mini) | Set 2 (4o-mini) | Set 1 (4o) | Set 2 (4o) |
|---|---|---|---|---|
| NaCl | 0.0 | 0.0 | 0.4249 | 0.4213 |
| KCl | 0.0 | 0.0 | 0.0392 | 0.0577 |
| NH4Cl | 0.0 | 0.0 | 0.7578 | 0.7952 |
| Na2SO4 | 0.0 | 0.0 | 0.569 | 0.5584 |
| K2SO4 | 0.0444 | 0.0836 | 0.3471 | 0.3051 |
| NH4NO3 | 0.0 | 0.0 | 0.5294 | 0.5362 |
| NaH2PO4 | 0.0 | 0.0 | 0.7905 | 0.8065 |
| NaNO3 | 0.0567 | 0.0391 | 0.8269 | 0.8488 |
| Na3PO4 | 0.2289 | 0.2266 | 0.8704 | 0.8756 |
| KBr | 0.0 | 0.0 | 0.0198 | 0.0769 |
| KNO3 | 0.2959 | 0.1818 | 0.8235 | 0.7879 |
| RbCl | 0.0 | 0.0 | 0.3051 | 0.2456 |
| Macro F1 | 0.0522 | 0.0443 | 0.5253 | 0.5263 |